\documentclass[letterpaper ,conference]{IEEEtran}
\IEEEoverridecommandlockouts
\usepackage{cite}
\usepackage{amsmath,amssymb,amsfonts}
\usepackage{algorithmic}
\usepackage{algorithm}
\usepackage{cleveref}
\usepackage{tabularx}
\usepackage{booktabs}
\usepackage{graphicx}
\usepackage{textcomp}
\usepackage{pgfplots}
\def\BibTeX{{\rm B\kern-.05em{\sc i\kern-.025em b}\kern-.08em
    T\kern-.1667em\lower.7ex\hbox{E}\kern-.125emX}}

\newcommand{\ourdata}{URSA }

\begin{document}

\title{Unlimited Road-scene Synthetic Annotation (URSA) Dataset}

\author{\IEEEauthorblockN{Matt Angus$^{\ddagger}$, Mohamed ElBalkini$^\dagger$, Samin Khan$^\dagger$, Ali Harakeh$^*$, \\ Oles Andrienko$^*$, Cody Reading$^*$,
\\ Steven Waslander$^*$, and Krzysztof Czarnecki$^\dagger$}
\IEEEauthorblockA{\textit{Computer Science}$^{\ddagger}$, \textit{Electrical and Computer Engineering$^\dagger$}, \textit{Mechanical and Mechatronics Engineering$^*$} \\
\textit{University of Waterloo}\\
Waterloo, ON, Canada \\
m2angus@uwaterloo.ca, melbalki@uwaterloo.ca, sa24khan@uwaterloo.ca, www.aharakeh.com,
\\ oandrien@uwaterloo.ca, careadin@uwaterloo.ca,
\\ stevenw@uwaterloo.ca, k2czarne@uwaterloo.ca}
}

\maketitle

\begin{abstract}
In training deep neural networks for semantic segmentation, the main limiting factor is the low amount of ground truth annotation data that is available in currently existing datasets. The limited availability of such data is due to the time cost and human effort required to accurately and consistently label real images on a pixel level. Modern sandbox video game engines provide open world environments where traffic and pedestrians behave in a pseudo-realistic manner. This caters well to the collection of a believable road-scene dataset. Utilizing open-source tools and resources found in single-player modding communities, we provide a method for persistent, ground truth, asset annotation of a game world. By collecting a synthetic dataset containing upwards of $1,000,000$ images, we demonstrate real-time, on-demand, ground truth data annotation capability of our method. Supplementing this synthetic data to Cityscapes dataset, we show that our data generation method provides qualitative as well as quantitative improvements---for training networks---over previous methods that use video games as surrogate.
\end{abstract}


\section{Introduction}
The main barrier for improving deep semantic segmentation models can be attributed to the small scale of existing semantic segmentation datasets, which range from only hundreds of images \cite{brostow2009semantic} to just a few thousand \cite{cordts2016cityscapes, Neuhold_2017_ICCV}. The widely used KITTI Benchmark Suite has $7,481$ training and $7,518$ testing images for object detection, but only $200$ training and $200$ testing images for semantic segmentation \cite{geiger2012we}. Why are semantic segmentation datasets so small? The answer lies in the time required to obtain high quality semantic segmentation annotation. For example, the Camvid dataset \cite{brostow2009semantic} reports requiring \textbf{60} minutes for annotating a single frame. For the much higher resolution Cityscapes and Mapillary Vistas datasets \cite{cordts2016cityscapes, Neuhold_2017_ICCV}, annotation time increases to \textbf{90-94} minutes per frame. The substantial time required to annotate a single frame is due to the difficulty of tracing accurate object boundaries. As the resolution increases, this problem becomes more prominent, making it prohibitive to have high resolution datasets larger than a few thousand frames for semantic segmentation.

As a workaround for the lack of large-scale datasets, semantic segmentation models \cite{badrinarayanan2015segnet,long2015fully} are usually initialized from image classifiers and then fine-tuned on the much smaller semantic segmentation datasets \cite{cordts2016cityscapes, brostow2009semantic, geiger2012we}. This is termed \textit{transfer learning} \cite{pan2010survey} and has been demonstrated to work well in practice. However, reusing parameters from a pre-trained image classification model constrains semantic segmentation model architectures to be similar to the one used in image classification. Even slight modification of models initialized from image classification weights is shown to produce optimization difficulties \cite{yosinski2014transferable}. As an alternative, and to enable more freedom in model architecture exploration, researchers have recently begun employing synthetic data \cite{gaidon2016virtual,ros2016synthia,richter2016playing} as a surrogate for real data when training semantic segmentation models. 
 
Currently, two prominent approaches are available to generate synthetic data for semantic segmentation. The first relies on virtual world generators created with development platforms such as Unity \cite{ros2016synthia,gaidon2016virtual} or Unreal \cite{mueller2017ue4sim}. Virtual worlds created by researchers using this approach are usually highly configurable and allow the collection of large amounts of data as no manual annotation is needed. Furthermore, a broad range of useful information can be generated from these worlds as the developer has full access to all the data within the generator. A drawback of this approach is the significant gap in quality of generated images with respect to real images. This causes domain shift \cite{Tommasi2016}---a difference in image generating distributions between the natural and synthetic images. Domain adaptation methods have been used to combat this phenomenon, but generating synthetic data that closely resembles real data is the most efficient way to minimize this difference. To that end, a second approach has begun to emerge, where researchers use video games such as Grand Theft Auto V (GTAV) to produce datasets of ultra-realistic scenes and corresponding ground truth annotations. These datasets have been created for a variety of tasks such as object detection \cite{johnson2017driving,richter2017playing}, and semantic segmentation \cite{richter2016playing,richter2017playing}. However, the internal operation and content of off-the-shelf games are largely inaccessible, making it difficult for researchers to get detailed annotations for semantic segmentation without human annotators. It has to be noted that the same virtual world generators used for open-source simulators are also used for video game creation. However, commercial video game makers have access to a much larger pool of resources---than small research teams---to enable production of more realistic-looking graphics and robust physics engines.

This paper explores using commercial video games to generate large-scale, high-fidelity training data for semantic segmentation with minimal human involvement. We overcome the inaccessibility of in-game content by leveraging tools traditionally used for game modifications. Specifically, utilizing tools from the single-player modding community of Grand Theft Auto V (GTAV), we create a texture pack that replaces in-game surface textures with solid colors belonging to $37$ classes (Table \ref{table:classcomp}) commonly used for semantic segmentation in the context of autonomous driving. The texture pack---created from human annotator input--- is coupled with a data collection mechanism to collect infinite amounts of semantic segmentation ground truth data at the cost of a single in-game assets annotation run. To summarize, our contributions are:
\begin{itemize}
\item We provide a method to generate unlimited amounts of semantic segmentation data from computer games at a constant human annotation time.
\item We provide a dataset that is $4$x larger than current state-of-the-art open source synthetic road-scene segmentation datasets and exceeds the training set size capabilities of current state-of-the-art segmentation networks.
\item We demonstrate that our proposed in-game permanent texture replacement modification generates more exhaustive and better quality labelling than the temporal label propagation employed in prior work \cite{richter2016playing,richter2017playing}.
\item We show that training neural networks on data generated by our method reduces the amount of real data---and the number of training iterations---required by such models to converge.
\end{itemize}

\section{Related Work} \label{related_work}
Although tedious and time consuming to construct, attempts to produce large semantic segmentation datasets continue unabated. The Camvid dataset \cite{brostow2009semantic} contains $701$ annotated images with $32$ labelled categories. Due to the limited size of this dataset, only the $11$ largest categories are typically used for training deep models. The data is mostly collected around Cambridge, UK, making it of limited use when trying to generalize to new scenes. The Cityscapes dataset \cite{cordts2016cityscapes}, collected in $50$ cities around Germany, Switzerland, and France is the largest, non-commercial, semantic segmentation dataset available for researchers. It has $5,000$ finely annotated images and $20,000$ coarsely annotated ones spanning a total of $30$ classes with only $19$ being used for evaluation. The annotation time for fine labels is $90$ minutes per image. The much larger, recently released Mapillary Vistas dataset \cite{Neuhold_2017_ICCV} contains $25,000$ finely annotated images spanning 66 classes. It uses the same annotation mechanism employed in \cite{cordts2016cityscapes}, averaging about $94$ minutes of annotation time per image. The dataset is commercial and only a portion is available for use free of charge. The large annotation time required for these datasets imposes a significant annotation budget that is not suitable for non-commercial entities. 

To that end, there has been a considerable increase in the use of synthetic data for training deep semantic segmentation models. For instance, the Synthia dataset \cite{ros2016synthia} contains $13,400$ annotated frames for the semantic segmentation task. Since the dataset is constructed from a virtual city implemented with the Unity development platform, more data can be collected at any time at zero additional annotator budget. Virtual KITTI \cite{gaidon2016virtual} is another dataset captured from a Unity based simulation environment, in which real-world video sequences are used as input to create virtual and realistic proxies of the real-world. The dataset is comprised of $5$ cloned worlds and $7$ variations of each summing up to  $35$ video sequences with roughly $17,000$ frames. The synthetic videos are coupled with ground truth annotations for RGB tracking, depth, optical flow, and scene segmentation. Both of these datasets share a common disadvantage, where the provided data is far from being considered realistic. Richter et al. \cite{richter2017playing} performed a perpetual experiment to determine the realism of common synthetic datasets. Workers on Amazon Mechanical Turk (AMT) were asked to determine which synthetic images they think are closer in appearance to the Cityscapes dataset. Their video game based dataset was shown to be more realistic than both the Synthia and Virtual KITTI datasets. To further fortify this conclusion, we show in Section \ref{exp} that deep models trained on only synthetic data from video games outperform those that are trained on data generated from the Synthia dataset when tested on real data from the Cityscapes dataset. 

Playing for Data (PFD) \cite{richter2016playing} was one of the first works to use video games for the generation of photo-realistic data for the semantic segmentation task. PFD relies on inserting middleware between the game engine and the graphics hardware to extract information from GTAV without having access to the game's code or content. This process, known as detouring, grants access to coarse-level, in-game information, which is then used to section the image into patches. Human annotators are then employed to label these patches for an initial image. Label propagation is then used to label consecutive frames with human annotators only intervening if label propagation does not annotate $97\%$ of a new frame. The labelling approach greatly reduces the annotation time required per frame, but still requires around $49$ annotator hours to label $25,000$ frames. Furthermore, the frames are restricted to be consecutive, limiting the variability of data collection in the game world. Richter et al. \cite{richter2017playing} expanded on their earlier work \cite{richter2016playing} to generate $250,000$ images with semantic segmentation labels. Although this method caches some of the information to disk to be reused, their annotation time still grows with the number of generated frames. This is due to the fact that there is no way to know if all assets in a scene are labeled, unless the scene has been rendered. If an asset was encountered for the first time, human annotators need to intervene to generate a label for that asset. Finally, detouring was also employed to access the resources used for rendering. Detouring can only access coarse level information from the video game, limiting the number of classes that occur in this dataset. However, our proposed method requires a constant annotator time of $426$ hours to generate \textit{unlimited} amounts of road scene semantic segmentation data. Furthermore, since we label the game at a texture level, we have access to more refined class categorization than both \cite{richter2016playing} and \cite{richter2017playing}. For more information on detouring, we refer the reader to \cite{richter2016playing}.

The two main benefits of simulation based approaches over video game based approaches are the ability to have more fine-grained annotations---such as lane separator---and persistence of labels throughout the simulation. The drawback of simulation based approaches is a lack of realism. However, the method presented to generate URSA maintains fine-grained annotations, shown in Table \ref{table:classcomp}, and persistence while leveraging realistic rendering techniques of the game engine.

\section{Technical Approach}
\label{tech}
Our proposed data generation scheme can be separated into three steps. First, we uniquely identify super-pixels in an image by correlating File path of a drawable, Model name, Shader index, and Sampler (FMSS) data that we parse using the open-source Codewalker tool \cite{codewalker}. The FMSS sectioning results in $1,178,355$ total in-game sections (hereafter refered to as FMSS), of which $56,540$ are relevant for generating our road dataset. FMSS that occur indoors, for example, are not considered. Second, we create a user interface to collect annotations for every relevant FMSS via AMT. Third, we create a data collection framework that exploits in-game AI to drive around and collect dashcam-style frames to be used to render the \ourdata dataset\footnote{The publisher of Grand Theft Auto V allows non-commercial use of footage from the game as long as certain conditions are met \cite{singleplayermods,rockstarpolicy}.}.

\subsection{Reducing Annotation Effort Via View Selection}

Labelling FMSS out of road scene context is very inefficient and difficult to do. Richter et al. \cite{richter2016playing, richter2017playing} provided a method to label super-pixels from GTAV in context, by creating a GUI where users label frames at the super-pixel level. Our task involves labeling all in-game FMSS and not just in-frame ones. One option to accomplish this task is to randomly choose frames until all FMSS in the game are labeled. However, this method provides no upper bound on the number of frames required to label all in-game FMSS. To that end, we devise a view selection mechanism that allows us to generate a near-optimal number of frames required to label the 56,540 FMSS. We exploit the paths used by the in-game AI drivers that form a graph $G = (V,E)$. Each vertex, along with position, contains information about the road type, which is used to choose only major roads for view selection (i.e. no dirt roads, alleyways). The edges define the structure of the road network. The goal is to find a minimal set $(M,L) \subseteq V \times E$ such that all FMSS used in outdoor scenes are existent in at least one scene. Here $M$ is the set of vertices to view from and $L$ is the edge to look down.

We partition $V$ into three sets
\begin{align*}
    V_{\text{simp}} &= \{ v | v \in V, \deg(v) = 2 \}\\
    V_{\text{comp}} &= \{ v | v \in V, \deg(v) > 2\}\\
    V_{\text{dead}} &= \{ v | v \in V, \deg(v) < 2\}.
\end{align*}
$V_{simp}$ consists of simple road sections, where finding the direction of travel is trivial. $V_{comp}$ consists of more complex interchanges such as merge lanes or intersections, whereas $V_{dead}$ consists of dead ends to be ignored. The direction of travel for $V_{comp}$ is determined by clustering these interchanges with DBscan \cite{ester1996density}---a density based spatial clustering algorithm---then applying linear regression to estimate the road direction. Determining a minimal set that includes all FMSS from this graph is NP-hard, so some desirable properties of $M$ and $L$ are developed. First, we require a minimum separation between views, $d_{\text{min}}$ in Eq. \ref{min_dist}, and second, that two consecutive view points look in the same direction in Eq. \ref{look_dir}. These two properties are summarized as

\begin{align}
    &\forall{u,v} \in M, ||u - v|| > d_{\text{min}} \label{min_dist}\\
    &\forall{u,v} \in M, \forall{u_1,v_1} \in \varphi(u,v),\label{look_dir}\\ \nonumber
    &\quad \ \  vv_1 \in E \land uu_1 \in E \Rightarrow uu_1 \in L \oplus vv_1 \in L ,
\end{align}
where $\varphi(u,v)$ denotes the the shortest path from $u$ to $v$, $d_{\text{min}}$ is a tunable parameter that controls the minimum distance between each selected vertex, and $\oplus$ is the exclusive or operator. The larger the value of $d_{\text{min}}$, the more likely two consecutive scenes miss FMSS. Eq. \ref{min_dist} allows us to approximate full coverage of road scenes. To prevent annotators from potentially mis-labelling far away objects, we set a maximum distance after which FMSS in the scene are ignored. Finally, viewpoint selection results in $3,388$ viewpoints having a coverage over the aforementioned $56,540$ FMSS. Without Eq. \ref{min_dist}, $M$ should equal $V$ and $L$ should equal $E$, making FMSS annotation unfeasible. Without Eq. \ref{look_dir}, $L$ would contain edges that point towards each other, leading to many redundant FMSS to annotate.

\subsection{Efficient FMSS Annotation}
Now that we have full game coverage for the required FMSS, we need to devise an efficient annotation mechanism in a way that allows for the recycling of labels. We inspect the game files for a unique identifier to address textures on disk rather than on the volatile graphics buffers previously used in \cite{richter2016playing}. We determine the unique identifier FMSS that allows us to inject textures when requested by the game engine using tools developed by the game modding community such as OpenIV \cite{OIV}. Since labelling textures on disk persist indefinitely, any number of frames can be rendered multiple times in either realistic or ground truth textures.

\begin{figure} 
\includegraphics[width=\columnwidth]{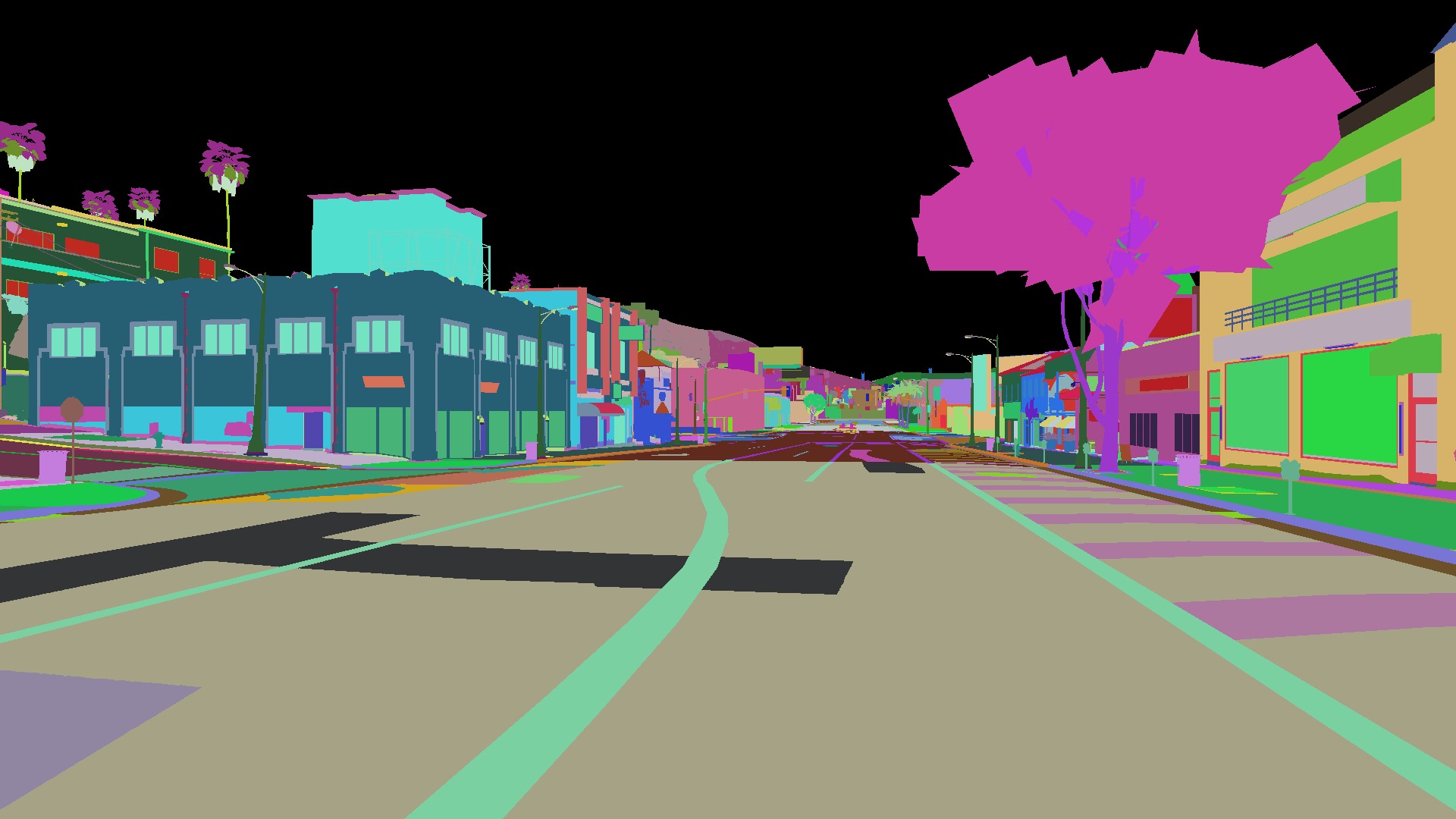}
\caption{Sections with unique FMSS tuple. Each FMSS tuple is assigned a random colour in this image.}
\label{mtsscolour}
\end{figure}

Similar to Richter et al. \cite{richter2017playing}, we ease the pixel labelling burden by identifying super-pixels that uniquely correlate to a portion of an in-game object. As mentioned previously this unique identifier is based on FMSS. Finding all FMSS in a rendered scene leads to the over segmented image shown in Figure \ref{mtsscolour}. However, each super-pixel in the scene may be influenced by multiple FMSS (e.g. partially transparent textures). To remedy this problem, we identify super-pixels in each scene by weighing the influence of different FMSS at a pixel-level. Pixel influence is calculated by magnitude of contribution an FMSS to a given pixel. Each pixel is then assigned to the FMSS tuple with the highest influence. For example, road damage textures are blended with the road texture, they will both influence a set a pixels thus the maximum influence of the two is chosen for each pixel.

\subsection{Label Collection GUI}\label{GUI}
\begin{figure}[t]
	\centering
	\includegraphics[width=\columnwidth]{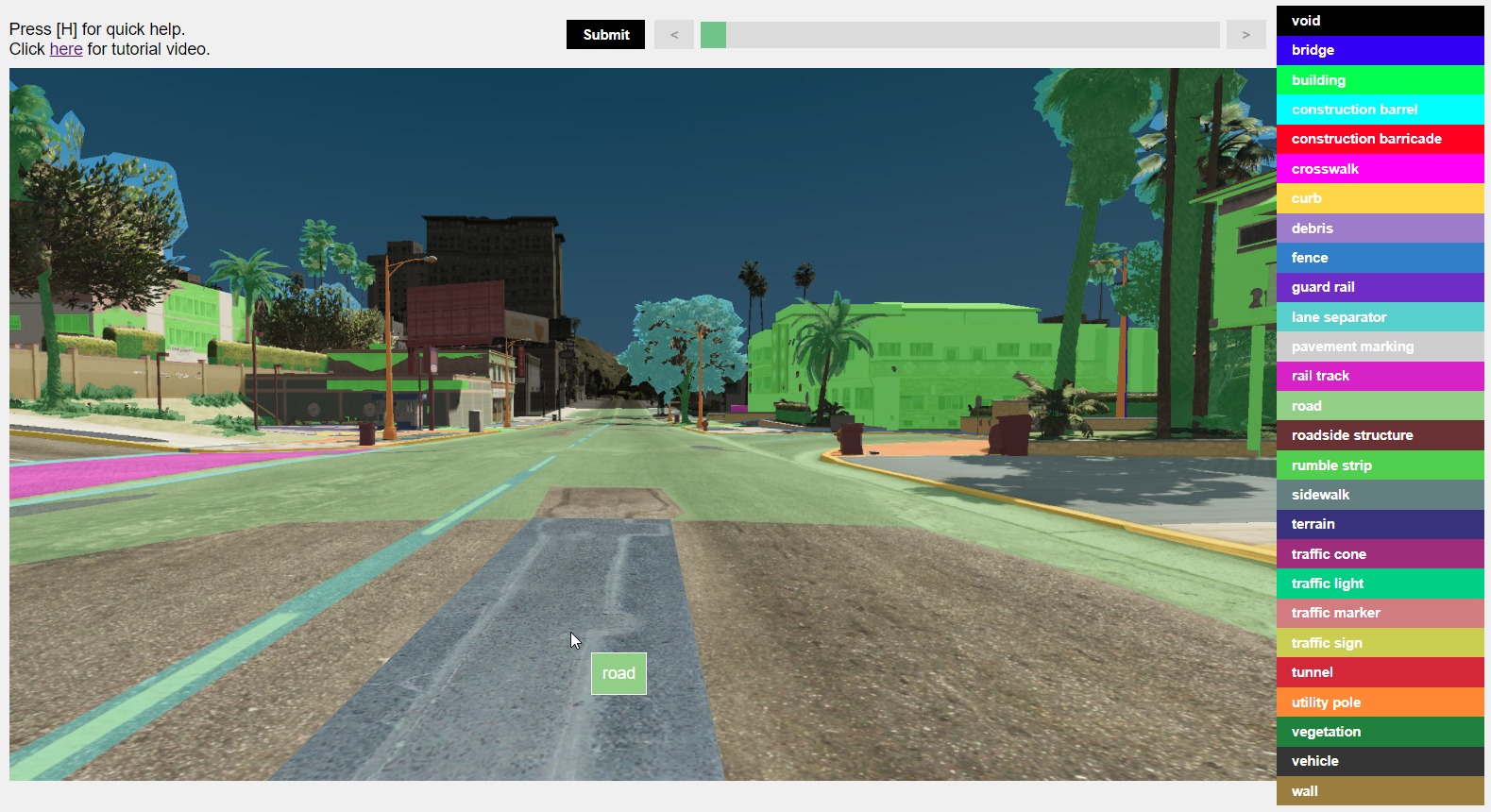}
	\caption{The Amazon Mechanical Turk user interface created for collecting texture annotations.}
	\label{fig:mturk_ui}
\end{figure}

The $3,388$ viewpoints are used to generate an equal number of dashcam-style scene snapshots, with each snapshot sectioned into FMSS as shown in Fig \ref{mtsscolour}. We design a web-based annotation GUI that allows annotators to classify each FMSS into one of the 28 road scene classes (Fig. \ref{fig:mturk_ui}).

Note that some of the 37 classes contained in our dataset (Table \ref{table:classcomp}) are not considered in this image-based annotation process. Dynamic objects---including around $500$ vehicle, $900$ pedestrian, and $30$ animal models---are annotated by domain experts to ensure the highest accuracy. These in-game assets demand a relatively small amount of annotation effort compared to static road-scene objects in GTAV.  For instance, the time required to exhaustively search for all relevant animal and pedestrian assets does not exceed 2-3 hours for a single annotator.

In order to gather labels for the static world objects, we expose the aforementioned GUI to AMT service. By leveraging the parallel efforts of numerous helpful AMT workers, we gather annotations for the static super-pixels representing FMSS. We evaluate the efficiency of our annotation method in section \ref{exp}.

\subsection{Data Generation}
To record the \ourdata dataset, we simulate a dashboard camera and record in-game scenes. We set the AI to drive the vehicle along main roads throughout the game world using a community mod called DeepGTAV \cite{deepgtav}. We record gameplay clips using Rockstar Editor's in-game recording feature \cite{rockstareditor}. Rockstar Editor is an offline rendering mode which removes the real-time constraint to maintain a high frame rate. We then use Rockstar Editor mode in order to render the recorded clips. First, we render the realistic frames using the maximum settings of the GTAV game engine without lens distortion and chromatic aberration effects. However, the same methodology cannot be used to render ground truth data. Shaders responsible for blending textures, casting shadows, and mimicking lens flare need to be replaced so that the ground truth textures are rendered without artifacts. To perform this task we create our own shaders that guarantees ground truth data to be rendered without artifacts. Afterwards, we re-render the clips using the replacement shaders and accumulated annotations from AMT workers to generate the semantic segmentation ground truth. An example of the ground truth frame and its realistic counterpart are shown in Fig. \ref{fig:qual_imgs}. We run two separate copies of the game with our automated Rockstar Editor rendering procedure on two separate PCs simultaneously. With this setup, we are able to record and render the entirety of the \ourdata dataset---1,355,568 images---with a total computation budget of around 63 hours.

\section{Experiments}\label{exp}
\subsection{Efficiency Of The Proposed Annotation Scheme}
We leverage the resources of Amazon Mechanical Turk (AMT) to gather annotations using the GUI presented in Section \ref{GUI}. We design AMT tasks such that each task contains $2-6$ images, with a maximum of $270$ FMSS to be annotated per task. We provide each AMT worker $20$ minutes to finish the task.

A preliminary performance analysis was done to determine how many votes are required to achieve a desired level of annotation accuracy. A random sample of $200$ segments was selected and labeled by domain experts. We then test AMT workers' performance against this expert-annotated subset of data to determine at what point there is diminishing returns for the number of votes. It can be seen in Fig. \ref{fig:mturk_cost_benefit} that diminishing returns is around 6-7 votes per FMSS, achieving around $75\%$ accuracy. This procedure was followed to minimize the cost while keeping the accuracy high for the task at hand.

Using this cost optimizing procedure, it took around $426$ AMT hours to label the $3,388$ scenes---about $22\%$ of the annotation time needed for Playing For Benchmarks (PFB) \cite{richter2017playing}. We achieved an average of $7.04$ votes per FMSS. This average only includes FMSS that appear in at most 11 scenes ($95\%$ of tuples)---including very common objects would skew this average. Furthermore, since we are annotating FMSS instead of per-frame super-pixels, we can recycle these throughout the game. This allows us to generate an unlimited amount of data without any performance drawbacks.
\begin{figure}[t]
\caption{AMT Sample Set Cost Benefit Correlation}\label{fig:mturk_cost_benefit}
\centering
\begin{tikzpicture}
\begin{axis}[
    xlabel={\# of votes},
    ylabel={\% correct labels},]
    xmin=0, xmax=20,
    ymin=0.2, ymax=1.1,
    xtick={0, 1, 2, 3, 4, 5, 6, 7, 8, 9, 10, 11, 12, 13, 14, 15, 16, 17, 18, 19},
    ytick={0.2, 0.3, 0.4, 0.5, 0.6, 0.7, 0.8, 0.9, 1.0},
    ymajorgrids=true,
    xmajorgrids=true,
    grid style=dashed,
]
\addplot+[
    color=blue,
    mark=.,error bars/.cd,
     y dir=both,y explicit
    ]
    coordinates {
    (1,0.2814) +- (0,0.021611108254784173)
    (2,0.3011398963730571) +- (0,0.020376335264050492)
    (3,0.48643356643356656) +- (0,0.0347168816427417)
    (4,0.5899259259259259) +- (0,0.028717822095320714)
    (5,0.7008264462809918) +- (0,0.026238855977500072)
    (6,0.7298876404494379) +- (0,0.025320568127646374)
    (7,0.7897674418604652) +- (0,0.029767441860465114)
    (8,0.7866666666666667) +- (0,0.029886617014028373)
    (9,0.7959322033898304) +- (0,0.0293489638263232)
    (10,0.8349999999999999) +- (0,0.029415604324538002)
    (11,0.8421052631578945) +- (0,0.023537557657892536)
    (12,0.9046153846153845) +- (0,0.024615384615384647)
    (13,0.883809523809524) +- (0,0.02363747361141113)
    (14,0.9266666666666665) +- (0,0.025915341754868013)
    (15,0.9015384615384617) +- (0,0.034538375877913335)
    (16,0.9818181818181819) +- (0,0.036363636363636376)
    (17,0.9879999999999999) +- (0,0.032496153618543834)
    (18,0.9955555555555555) +- (0,0.02177324215807271)
    (19,0.9942857142857143) +- (0,0.027994168488950585)
    };
    
\end{axis}
\end{tikzpicture}
\end{figure}
\subsection{Quality of Collected Data:}

\noindent\textbf{Baselines and Evaluation Metrics:}
We compare our generated data to that of Playing For Benchmarks (PFB) \cite{richter2017playing} and the Synthia datasets \cite{ros2016synthia}. We generate around $1,355,568$ frames of both realistic images and semantic segmentation ground truth labels to serve as our dataset. For training and testing on a real dataset, we choose the Cityscapes dataset \cite{cordts2016cityscapes}, which includes $3,475$ total frames split into $2,975$ frames for training and $500$ frames for validation. We use class-specific Intersection-Over-Union (c-IOU) to measure the performance of semantic segmentation algorithms as described by \cite{cordts2016cityscapes}.\\

\noindent\textbf{Neural Network Baselines:} 
We choose two neural network architectures, SegNet \cite{badrinarayanan2015segnet} and FCN \cite{hoffman2016fcn}, as our semantic segmentation algorithms. We use the default learning rate schedule provided by authors for all experiments. We use a batch size of 2 for Segnet and 1 for FCN as recommended by the authors.\\

\begin{figure*} [t]
    \centering
    \includegraphics[width=\textwidth]{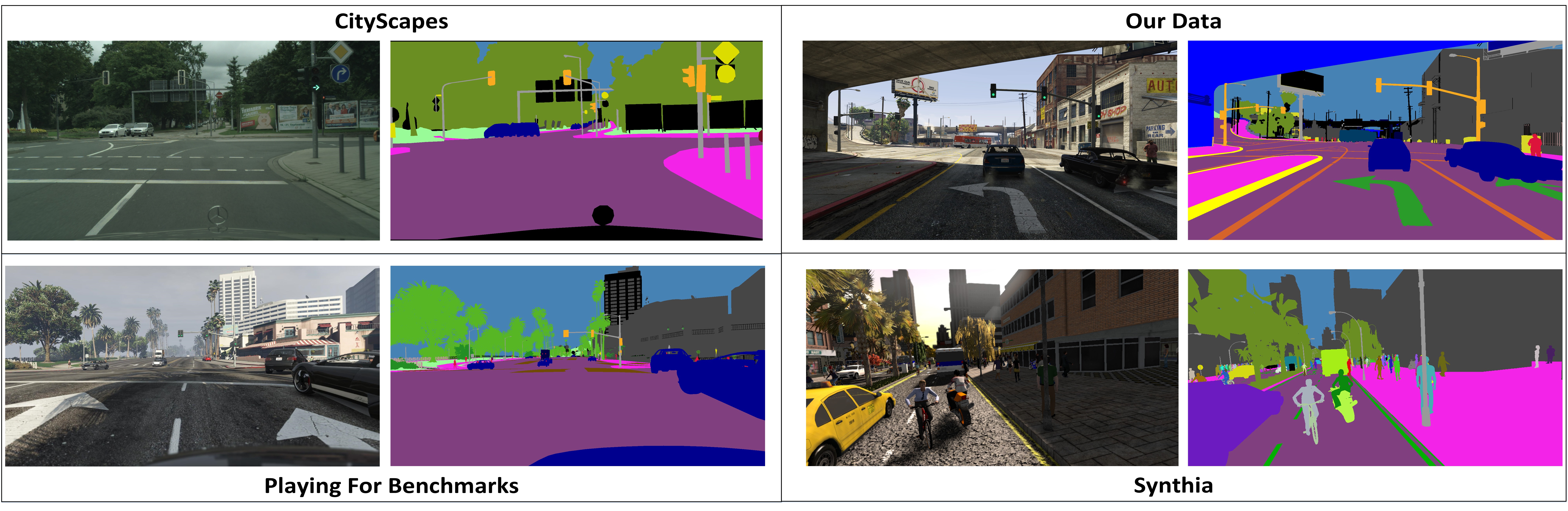}
    \caption{Comparison of realistic (left) and ground truth (right) images.}
    \label{fig:qual_imgs}
\end{figure*}

\noindent\textbf{Experiments and Results:}
To validate the quality of our generated data in comparison to other synthetic datasets, we devise several experiments involving training two deep neural network models. We divide our experiments to measure two characteristics of synthetic datasets:
\begin{itemize}
    \item \textbf{Closeness To Real Data}: Here we train the neural networks for $110K$ iterations on each synthetic dataset with VGG-16 ImageNet weight initialization. We refer to this experiment as ``Train Only".
    \item \textbf{Reduction In The Amount of Required Real Data:} Here we train for $100K$ iterations on synthetic datasets with VGG16 ImageNet initialization, then fine-tune for 10K iterations on Cityscapes. Fine-tuning is performed on $25\%, 50\%, 75\%$, and $100\%$ of the $2,975$ training frames from the training set of Cityscapes. We refer to these experiments as ``FT-25\%", ``FT-50\%", ``FT-75\%", and ``FT-100\%" respectively.
\end{itemize}
The resulting neural networks from all experiments are tested on the $500$ frames from the Cityscapes validation data split. The ``Train Only" experiment provides us with a quantitative assessment of the closeness of the data generating distribution between the synthetic and the real datasets. The ``FT-25, -50, -75, -100" experiments allow us to quantitatively show which synthetic dataset allows a greater reduction in the amount of real data needed to train neural networks while maintaining performance. For consistency, we remap the classes available in each dataset to that of Cityscapes, as training on different classes in different datasets will result in a biased comparison. Since our data as well as PFB have more than the $220K$ frames required to perform the experiments, we randomly shuffle and choose frames from these datasets to get our required number. For comparison, we provide the results of training both deep models on all of the training data in the Cityscapes dataset for $110K$ iterations under ``Train Only, CS" experiment. \\

\noindent\textbf{Quantitative Results:}
Tables \ref{table:FCN} and \ref{table:SegNet} provides quantitative results of the predicted semantic segmentation labels measured through c-IOU, averaged over all 19 classes used in evaluation of Cityscapes. Furthermore, we present the results of the c-IOU a subset of classes that are directly related to \textit{free space estimation}, and \textit{vehicle detection}.

\begin{table*}[t]
\centering
\caption{Results of experiments on Cityscapse (CS), Synthia (SY), Playing for Benchmarks (PFB) and \ourdata using FCN \cite{long2015fully} as the baseline neural network.}
\label{fcn_results}
\resizebox{\textwidth}{!}{
\begin{tabular}{c||c || ccc||ccc||ccc||ccc||ccc} \toprule
Experiment & \multicolumn{4}{c|}{Train Only}  & \multicolumn{3}{c|}{FT-25\%} & \multicolumn{3}{c|}{FT-50\%} & \multicolumn{3}{c|}{FT-75\%} & \multicolumn{3}{c}{FT-100\%} \\ \midrule
Dataset & CS & SY & PFB & \ourdata & SY & PFB & \ourdata & SY & PFB & \ourdata & SY & PFB & \ourdata & SY & PFB & \ourdata \\ \midrule
Road & 0.866 & 0.023 & 0.214 & \textbf{0.705} & 0.728 & 0.875 & \textbf{0.902} & 0.807 & 0.893 & \textbf{0.915} & 0.812 & 0.882 & \textbf{0.917} & 0.814 & 0.882 & \textbf{0.911}\\
Sidewalk & 0.6 & 0.112 & \textbf{0.216} & 0.112 & 0.5 & \textbf{0.558} & 0.538 & 0.526 & 0.558 & \textbf{0.564} & 0.536 & \textbf{0.585} & 0.580 & 0.538 & \textbf{0.599} & 0.586\\
Fence & 0.211 & 0 & \textbf{0.047} & 0.017 & 0.172 & 0.158 & \textbf{0.183} & 0.243 & 0.222 & \textbf{0.246} & 0.249 & 0.215 & \textbf{0.263} & 0.252 & 0.239 & \textbf{0.279}\\
Car & 0.832 & 0.408 & \textbf{0.56} & 0.429 & 0.767 & \textbf{0.808} & 0.798 & 0.785 & \textbf{0.812} & 0.810 & 0.795 & \textbf{0.823} & 0.820 & 0.799 & \textbf{0.829} & 0.823\\
Truck & 0.213 & 0 & \textbf{0.065} & 0.011 & 0.128 & 0.195 & \textbf{0.224} & 0.136 & 0.22 & \textbf{0.242} & 0.153 & 0.248 & \textbf{0.266} & 0.154 & 0.26 & \textbf{0.289}\\
\midrule
Mean c-IOU (19 Classes) & 0.449 & 0.126 & \textbf{0.170} & 0.139 & 0.372 & 0.411 & \textbf{0.422}  & 0.395 & 0.432 & \textbf{0.439} & 0.400 &0.439 & \textbf{0.447} &0.408 &0.444 & \textbf{0.449} \\  
\bottomrule
\end{tabular}
}
\label{table:FCN}
\end{table*}

\begin{table*}[t]
\centering
\caption{Results of experiments on Cityscapse (CS), Synthia (SY), Playing for Benchmarks (PFB) and \ourdata using SegNet \cite{badrinarayanan2015segnet} as the baseline neural network.}
\label{segnet_results}
\resizebox{\textwidth}{!}{
\begin{tabular}{c||c ||ccc||ccc||ccc||ccc||ccc} \toprule
Experiment & \multicolumn{4}{c||}{Train Only} & \multicolumn{3}{c||}{FT-25\%} & \multicolumn{3}{c||}{FT-50\%} & \multicolumn{3}{c||}{FT-75\%} & \multicolumn{3}{c}{FT-100\%} \\ \midrule
Dataset & CS & SY & PFB & \ourdata & SY & PFB & \ourdata & SY & PFB & \ourdata & SY & PFB & \ourdata & SY & PFB & \ourdata \\ \midrule
Road & 0.833 & 0.009 & \textbf{0.065} & 0 & 0.739 & 0.650 & \textbf{0.815} & 0.676 & 0.694 & \textbf{0.759} & 0.610 & 0.604 & \textbf{0.684} & 0.641 & 0.659 & \textbf{0.695}\\
Sidewalk & 0.464 & 0.034 & \textbf{0.071} & 0.004 & 0.403 & 0.372 & \textbf{0.447} & 0.347 & 0.351 & \textbf{0.443} & 0.325 & 0.298 & \textbf{0.422} & 0.334 & 0.327 & \textbf{0.429}\\
Fence & 0.170 & 0 & \textbf{0.087} & 0.007 & 0.154 & \textbf{0.176} & 0.109 & 0.166 & \textbf{0.206} & 0.110 & 0.190 & \textbf{0.206} & 0.117 & 0.185 & \textbf{0.213} & 0.129\\
Car & 0.723 & \textbf{0.238} & 0.191 & 0.199 & 0.635 & 0.675 & \textbf{0.684} & 0.681 & \textbf{0.692} & 0.675 & \textbf{0.695} & 0.642 & 0.694 & 0.638 & 0.668 & \textbf{0.700}\\
Truck & 0.075 & 0 & \textbf{0.004} & \textbf{0.004} & 0.037 & 0.041 & \textbf{0.049} & 0.005 & \textbf{0.07} & 0.056 & 0.007 & \textbf{0.063} & 0.056 & 0.009 & \textbf{0.079} & 0.071\\
\midrule
Mean c-IOU (19 Classes) & 0.412 & 0.109  &\textbf{0.132}  & 0.066  &0.357  &0.349  & \textbf{0.370} &0.357 &\textbf{0.367} & \textbf{0.367}  &0.362  &0.351 & \textbf{0.382}  & 0.360 &0.366 & \textbf{0.388} \\  
\bottomrule
\end{tabular}
}
\label{table:SegNet}
\end{table*}

\noindent \textit{Train Only Results:}  When we train FCN on URSA data and test on Cityscapes, we can observe that we get a 0.705 class IOU for the road class, a 0.491 increase over the IOU achieved by similar training on PFB. Furthermore, by just training on URSA, FCN was able to achieve a road class c-IOU only 0.161 less than that of an FCN trained purely on real data. This phenomenon extends to classes that comprise most of the scenes in the game such as cars and buildings, which implies that by focusing our data collection on scenes with a high number of pixels belonging to our target classes, closes the gap in performance between real and synthetic data. When comparing FCN trained with our data to that trained on other datasets, we can see that PFB dominates most of the c-IOU. We attribute this phenomenon to PFB being well balanced when it comes to number of pixels per class in the dataset. Since we have the ability to generate unlimited data, class balancing can be performed in future sets of data.

SegNet, on the other hand, seems to have difficulties transferring across domains, as all three SegNet networks trained only on the synthetic datasets had a mean c-IOU $< 15\%$. These results on SegNet show that domain adaptation is not just correlated to the quality of data, but also to the neural network architecture itself.
\\
\noindent \textit{Fine Tuning Results:} Fine tuning on the target data is a very simple way to adapt to the target domain. Pretraining both architectures on the URSA dataset leads to a $1-2\%$ increase in mean c-IOU over PFB, and a $4-5\%$ increase in the same metric over Synthia. Although the quality of images in the URSA dataset is similar to that of PFB, URSA contains frames that are separated by a minimum distance. This increases independence among member within a single minibatch, generating better gradient estimate and allowing for more optimal pretrained weights. 

The results of FT-$25\%$ experiments are quite surprising. Fine tuning both architectures was capable of closing the gap between the performance on the real vs synthetic data. However, to our surprise, FCN trained on data from GTAV (both ours and PFB) then finetuned for only $10$K iterations on $25\%$ of Cityscapes was found to surpass the one trained purely on Cityscapes on the road, fence and truck classes. These results get better as we increase the percentage of real data to be used for fine tuning.

With FT-$25\%$, SegNet pre-trained on our data was capable of outperforming PFB and Synthia on the road, sidewalk, and car classes, all of which are crucial for autonomous driving systems. However, unlike FCN, the performance of fine tuning SegNet decreases on some classes, while increasing on others, as the percentage of real data it is trained on increases. This phenomenon can be attributed to the low capacity of the SegNet model, forcing the model to choose its weights to provide a balanced performance over all classes. In this case, balance is reflected as a drop in performance over high c-IOU classes in order to bump up low c-IOU ones as the amount of data belonging to the latter increases. As expected, networks trained on Synthia fall behind ones trained on PFB and the USRA datasets in terms of c-IOU.

From our experiments, we can show that using any synthetic data to pre-train neural networks results in a sizable reduction in the amount of real data required for semantic segmentation. The few percent difference in c-IOU between FT-$25\%$ and FT-$100\%$ is not as significant as the time required to label $75\%$ of the Cityscapes dataset---around $3,495$ annotator hours. From the performed experiments, pre-training on synthetic data in general helps with getting a stable initial set of neural network parameters for the Semantic Segmentation task.
We also assert our hypothesis that video game data is closer to real data than open source simulators. Finally, these results highlight the importance of our data generation method, as we close the gap between open source simulators and constant size video game datasets by granting researchers the ability to generate as much data as new segmentation architectures require. These capabilities are most useful in designing new segmentation architectures that cannot be initialized from ImageNet classification weights. \\

\noindent\textbf{Qualitative Analysis of Our Data:}
While detouring \cite{richter2016playing, richter2017playing} has proven to be useful for extracting data from video games, it suffers from limitations that are mitigated by our approach. These limitations are not evident from the quantitative analysis section alone, as testing on Cityscapes classes does not provide adequate insight on their origin. The biggest issue is that most modern game engines are using deferred shading techniques, meaning that transparent objects are rendered separate from the diffuse pass. This characteristic limits extracting such objects via detouring. For example vehicle glass, an essential part of a vehicle was not labeled by Playing For Benchmarks.

A second artifact resulting from detouring is loss of semantics of the objects being rendered. The graphics API receives geometry to draw without any details about its origin. This ambiguity is not present in our scheme and as such, we can easily label or even remove massive amounts of objects based on dataset requirements. As an example, we are able to remove on-road ``trash" from the rendered ground truth entirely without the need for human annotators. Finally, detouring cannot produce ground truth frames in real-time. This is due to the fact that detouring requires continuous hash table look-ups to check for any unlabeled resources in the scene.

A third consequence of detouring, is the lack of fine class semantics. Table \ref{table:classcomp} shows the union of $49$ classes from Cityscapes, Synthia, Playing for Benchmarks, and \ourdata. It can be seen that our synthetic data captures 37 out of the 49 classes most of which are essential for autonomous vehicles. Furthermore, our data retains the fine semantic categorization provided by custom designed simulators such as Synthia, while retaining the hyper-realistic image quality provided by video games. This is manifested as the ability of URSA to include classes such as ``Pavement Marking", ``Traffic Marker", and  ``Curb", all of which are not included in PFB.

Fig. \ref{fig:qual_imgs} shows the output semantic segmentation ground truth frames from the two baselines in comparison to our method. Unlike PFB, our segmentation ground truth contains glass as part of vehicles and buildings.

\begin{table}[]
    \centering
    \caption{ A comparison of classes represented in Cityscapes, Synthia, PFB, and URSA}
    \label{table:classcomp}
    \resizebox{!}{0.47\textheight}{
    \begin{tabular}{c||c|c|c|c} 
Class & CS & SY & PFB & \ourdata \\ \midrule
Airplane &  &  & \checkmark & \checkmark \\ \midrule
Animal &  &  & \checkmark & \checkmark \\ \midrule
Bicycle & \checkmark & \checkmark & \checkmark & \checkmark \\ \midrule
Billboard &  &  & \checkmark &  \\ \midrule
Boat &  &  & \checkmark & \checkmark \\ \midrule
Bridge & \checkmark &  &  & \checkmark \\ \midrule
Building & \checkmark & \checkmark & \checkmark & \checkmark \\ \midrule
Bus & \checkmark & \checkmark & \checkmark & \checkmark \\ \midrule
Car & \checkmark & \checkmark & \checkmark & \checkmark \\ \midrule
Chair &  &  & \checkmark &  \\ \midrule
Construction Barrel &  &  &  & \checkmark \\ \midrule
Construction Barricade &  &  &  & \checkmark \\ \midrule
Crosswalk &  &  &  & \checkmark \\ \midrule
Curb &  &  &  & \checkmark \\ \midrule
Debris &  &  &  & \checkmark \\ \midrule
Fence & \checkmark & \checkmark & \checkmark & \checkmark \\ \midrule
Fire Hydrant &  &  & \checkmark &  \\ \midrule
Guardrail & \checkmark &  &  & \checkmark \\ \midrule
Infrastructure &  &  & \checkmark &  \\ \midrule
Lane Separator &  & \checkmark &  & \checkmark \\ \midrule
Mobile Barrier &  &  & \checkmark &  \\ \midrule
Motorbike & \checkmark & \checkmark & \checkmark & \checkmark \\ \midrule
Parking Slot & \checkmark & \checkmark &  &  \\ \midrule
Pavement Marking &  &  &  & \checkmark \\ \midrule
Pedestrian & \checkmark & \checkmark & \checkmark & \checkmark \\ \midrule
Pole & \checkmark & \checkmark &  & \checkmark \\ \midrule
Rail Track & \checkmark &  & \checkmark & \checkmark \\ \midrule
Rider & \checkmark & \checkmark &  &  \\ \midrule
Road & \checkmark & \checkmark & \checkmark & \checkmark \\ \midrule
Roadside Structure &  &  &  & \checkmark \\ \midrule
Roadwork &  & \checkmark &  &  \\ \midrule
Rumble Strip &  &  &  & \checkmark \\ \midrule
Sidewalk & \checkmark & \checkmark & \checkmark & \checkmark \\ \midrule
Sky & \checkmark & \checkmark & \checkmark & \checkmark \\ \midrule
Terrain & \checkmark & \checkmark & \checkmark & \checkmark \\ \midrule
Traffic Cone &  &  &  & \checkmark \\ \midrule
Traffic Light & \checkmark & \checkmark & \checkmark & \checkmark \\ \midrule
Traffic Marker &  &  &  & \checkmark \\ \midrule
Traffic Sign & \checkmark & \checkmark & \checkmark & \checkmark \\ \midrule
Trailer & \checkmark &  & \checkmark & \checkmark \\ \midrule
Train & \checkmark & \checkmark & \checkmark & \checkmark \\ \midrule
Trash &  &  & \checkmark &  \\ \midrule
Trashcan &  &  & \checkmark &  \\ \midrule
Tree &  &  & \checkmark &  \\ \midrule
Truck & \checkmark & \checkmark & \checkmark & \checkmark \\ \midrule
Tunnel & \checkmark &  &  & \checkmark \\ \midrule
Van &  &  & \checkmark &  \\ \midrule
Vegetation & \checkmark & \checkmark & \checkmark & \checkmark \\ \midrule
Wall & \checkmark & \checkmark &  & \checkmark \\ \midrule
Total & 25/49 & 22/49 & 30/49 & \textbf{37/49} \\ \bottomrule
\end{tabular}
}
\end{table}

\section{Conclusion}
This paper has shown that video games can be used for large scale data generation using many techniques. The data that is produced is of greater quality than open source simulators, due to photo realistic rendering. However, simulators typically have more fine-grained annotations than video games. The method presented retains both of these benefits, as well as the ability to generate the same amount of data can be generated as open source simulators. We show that our dataset is superior to that of previous methods of using video games for training semantic segmentation models, especially the with road, sidewalk, and fence classes. A qualitative analysis for these improvements shows that ambiguity in graphics buffers and differed shading represents a major shortcomings of previous methods---mainly removal of artifacts, and real time rendering--that this work addresses.

{\small
\bibliographystyle{unsrt}

}
\end{document}